\documentclass[letterpaper]{article} 
\usepackage[draft]{aaai25}
\usepackage{times}  
\usepackage{helvet}  
\usepackage{courier}  
\usepackage[hyphens]{url}  
\usepackage{graphicx} 
\usepackage{natbib}  
\usepackage{caption} 
\usepackage{algorithm}
\usepackage{algorithmic}
\usepackage{amsmath,amsfonts}
\usepackage{multirow}
\usepackage{booktabs}
\usepackage{xcolor}
\usepackage{afterpage}
\usepackage{marvosym}
\usepackage{color}
\usepackage{marvosym}

\newcommand{\cww}[1]{\textcolor{red}{#1}}

\usepackage{color}
\definecolor{url_color}{RGB}{42, 83, 163}
\usepackage{newfloat}
\usepackage{listings}
\DeclareCaptionStyle{ruled}{labelfont=normalfont,labelsep=colon,strut=off} 
\floatstyle{ruled}
\newfloat{listing}{tb}{lst}{}
\floatname{listing}{Listing}
\title{DynaSurfGS: Dynamic Surface Reconstruction with \\ Planar-based Gaussian Splatting}

\author {
    Weiwei Cai\textsuperscript{\rm 1},
    Weicai Ye\textsuperscript{\rm 2,3,\textrm{\Letter}},
    Peng Ye\textsuperscript{\rm 3},
    Tong He\textsuperscript{\rm 3},
    Tao Chen\textsuperscript{\rm 1,\textrm{\Letter}}
}
\affiliations {
    \textsuperscript{\rm 1}Fudan University
    \textsuperscript{\rm 2}State Key Lab of CAD\&CG, Zhejiang University
    \textsuperscript{\rm 3}Shanghai AI Laboratory
    \\
  \texttt{\small \urlstyle{tt}\textcolor{url_color}{\url{https://open3dvlab.github.io/DynaSurfGS/}}}
    
}
\usepackage{bibentry}

\begin{document}

\twocolumn[{%
    \renewcommand\twocolumn[1][]{#1}%
    \setlength{\tabcolsep}{0.0mm} 
    \maketitle
    \begin{center}
        \newcommand{\teaserwidth}{\textwidth}
    \vspace{-4em}
        \includegraphics[width=\linewidth]{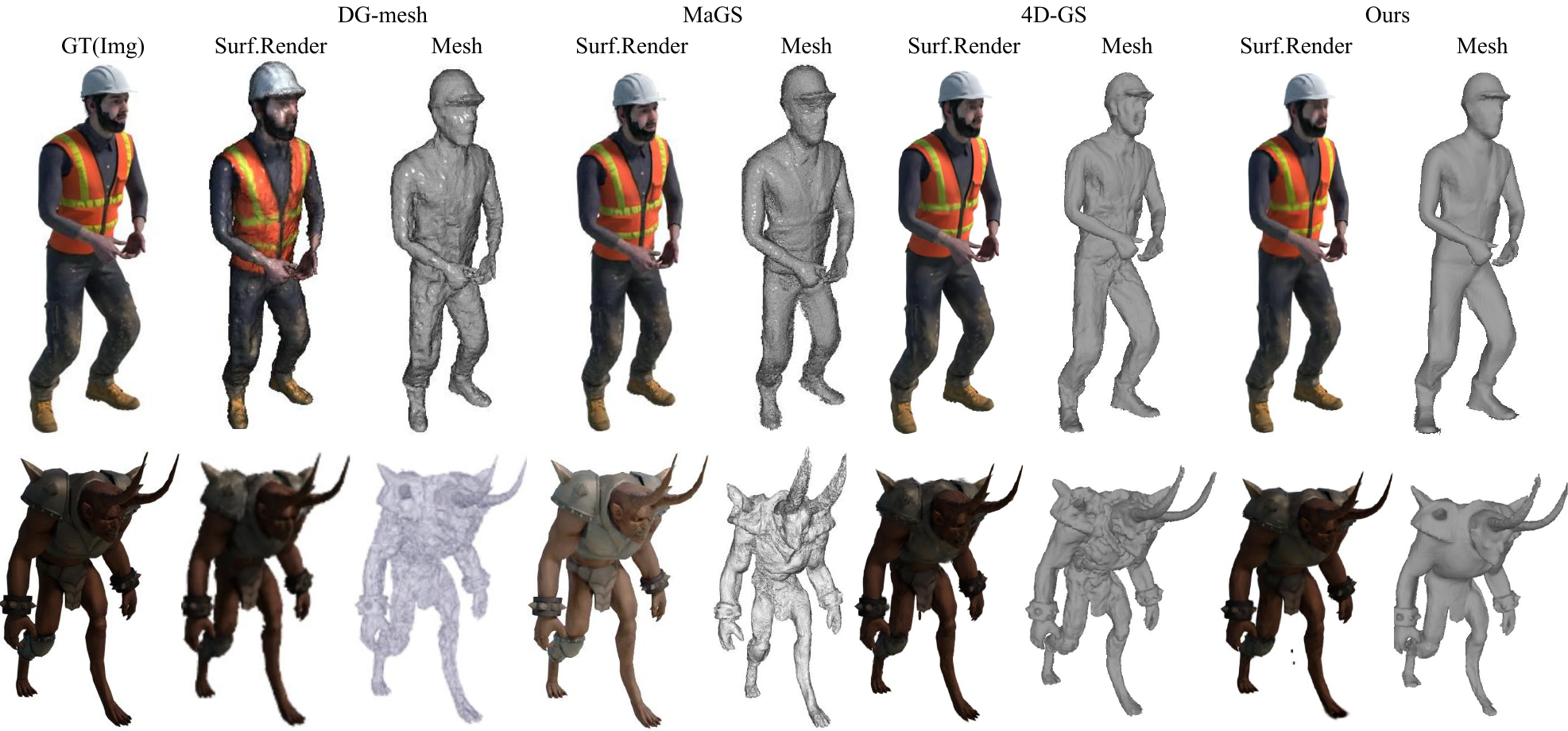}
      \vspace{-0.5cm}
        \captionof{figure}{We propose the DynaSurfGS framework, which can facilitate real-time photorealistic rendering and dynamic high-fidelity surface reconstruction. Compared with recent SOTA methods, such as DG-mesh~\cite{liu2024dynamic}, MaGS~\cite{ma2024reconstructing}, and 4D-GS~\cite{wu20244d}, DynaSurfGS achieves smooth surfaces with meticulous geometry.}
    \label{fig:teaser}
    \end{center}
}]
\begin{abstract}
Dynamic scene reconstruction has garnered significant attention in recent years due to its capabilities in high-quality and real-time rendering. Among various methodologies, constructing a 4D spatial-temporal representation, such as 4D-GS, has gained popularity for its high-quality rendered images. However, these methods often produce suboptimal surfaces, as the discrete 3D Gaussian point clouds fail to align with the object's surface precisely. To address this problem, we propose DynaSurfGS to achieve both photorealistic rendering and high-fidelity surface reconstruction of dynamic scenarios. Specifically, the DynaSurfGS framework first incorporates Gaussian features from 4D neural voxels
with the planar-based Gaussian Splatting to facilitate precise surface reconstruction. It leverages normal regularization to enforce the smoothness of the surface of dynamic objects.
It also incorporates the as-rigid-as-possible (ARAP) constraint to maintain the approximate rigidity of local neighborhoods of 3D Gaussians between timesteps and ensure that adjacent 3D Gaussians remain closely aligned throughout. Extensive experiments demonstrate that DynaSurfGS surpasses state-of-the-art methods in both high-fidelity surface reconstruction and photorealistic rendering. 

\end{abstract}
\section{Introduction}
Dynamic scene reconstruction from sparse data is a challenging and important task in computer vision, with a wide range of applications in fields such as movie production, entertainment industries, and autonomous driving. 
Many research efforts have been devoted to this field.
4D-GS~\cite{wu20244d} employs a deformation field network to model Gaussian motions and shape transformations, achieving remarkable rendering quality. SC-GS~\cite{huang2024sc} applies local weight interaction of sparse control points to obtain a 3D Gaussian motion field for high-quality rendering and editing.~\cite{lu20243d} integrates 3D geometric features into 3D deformation for efficient rendering and accurate representation of dynamic views. However, these approaches primarily focus on the rendering quality of dynamic synthesis while overlooking the geometry surface reconstruction of dynamic objects. Recently, some works have attempted to make uniform and improved mesh optimization. 
DG-Mesh~\cite{liu2024dynamic} jointly optimizes 3D Gaussian points and the corresponding mesh. MaGS~\cite{ma2024reconstructing} improves 3DGS by constraining Gaussian points to the mesh surface and modeling the relationship between the mesh and Gaussian function through a learnable deformation field. However, the reconstructed geometric surfaces produced by these methods often lack smoothness and remain unsatisfactory, as shown in Fig.~\ref{fig:teaser}. In short, achieving high-quality rendered images while simultaneously constructing satisfactory geometric surfaces of dynamic scenarios remains largely unexplored.

Although existing methods have achieved significant success in dynamic view synthesis and static geometry reconstruction, they still struggle to reconstruct smooth geometric surfaces while producing high-quality images of dynamic scenarios. For example, while 4D-GS~\cite{wu20244d} excels in rendering high-quality images, the resulting geometric surfaces are rough, as illustrated in Fig. \ref{fig:dnerf}. Traditional rasterization methods, such as 4D-GS, fail to accurately model the geometry of dynamic scenes because Gaussians, due to their disordered and irregular nature, do not adhere well to the surface of the scene. Similarly, PGSR~\cite{chen2024pgsr} performs well in reconstructing high-accuracy static object geometry but falls short in reconstructing dynamic objects, as shown in Fig. \ref{fig:dnerf}, owing to the absence of temporal information. Therefore, the research focuses on achieving high-quality rendered images while simultaneously constructing satisfactory geometric surfaces of dynamic scenarios.

In this paper, we introduce the DynaSurfGS framework, aiming to achieve high-quality rendered images alongside high-precision dynamic surface reconstruction. To this end, the 4D voxels are decomposed into 2D planes to encode the motion and shape changes of each Gaussian. A compact MLP is then employed to predict the deformation of the Gaussian across different timesteps. Unlike traditional rasterisation~\cite{kerbl20233d},  we calculate depth using a planar-based depth rendering approach inspired by PGSR \cite{chen2024pgsr}. This enables optimized Gaussian points to be attached to the scene surface. Subsequently, we employ planar-based Gaussian splitting to generate normal and distance maps, which are then utilized to compute an unbiased depth map, as shown in Fig. \ref{fig:framework}. We further estimate the distance normal using the unbiased depth information from neighboring pixels. By enforcing consistency between the distance normal and the rendering normal, we ensure that the depth aligns with the normal, facilitating the reconstruction of smooth geometric surfaces for dynamic objects.
To maintain the rigidity of dynamic objects during motion, we introduce the as-rigid-as-possible (ARAP) regularization, which constrains the object's shape to remain unchanged following rotation and translation.

We conducted comprehensive experiments on the D-NeRF~\cite{pumarola2021d} DG-Mesh~\cite{liu2024dynamic} and Ub4D ~\cite{johnson2023unbiased} dataset. Additionally, we perform extensive ablation studies to validate the effectiveness of our design. Our contributions are summarized as:
 \begin{itemize}
\item[$\bullet$] We propose the novel DynaSurfGS framework, which integrates the Hex-Plane representation and planar-based Gaussian splatting techniques to achieve photorealistic rendered images while enabling precise dynamic surface reconstruction.
\item[$\bullet$] We introduce normal regularization to constrain the reconstruction of geometric surfaces and apply the as-rigid-as-possible (ARAP) regularization 
to ensure consistency in the motion of dynamic objects. 
\item[$\bullet$] Extensive experiments demonstrate the effectiveness of the proposed DynaSurfGS framework. 
\end{itemize}
\section{Related Work}
\subsection{Dynamic View Synthesis}
Dynamic View Synthesis (DVS) aims to render novel photorealistic views from arbitrary viewpoints using monocular video of a dynamic scene at any given time step. DVS approaches can be broadly categorized into two main types: one involves modeling the dynamic scene using a learned deformation field that maps coordinates from each input image to a canonical template coordinate space. D-NeRF~\cite{pumarola2021d} is the first to combine the canonical field with the deformation field to effectively model dynamic scenes. Nerfies~\cite{park2021nerfies} applies a coarse-to-fine regularization to adjust the capacity of the deformation field to capture high frequencies during optimization. HyperNeRF~\cite{park2021hypernerf} extends the canonical field into a higher dimensional space to represent the 5D radiance field. 
Recently, a novel 3D scene representation using 3D Gaussian splitting has been proposed for dynamic view synthesis. Dynamic 3D Gaussians~\cite{luiten2023dynamic} is the first to adopt 3D Gaussians for dynamic view synthesis. ~\cite{yang2024deformable} decomposes dynamic scene into 3D Gaussians and a deformation field.~\cite{lu20243d} integrates 3D geometry-aware features into deformation fields to facilitate 3D Gaussian deformation learning.~\cite{huang2024sc} represent 3D dynamic scenes by manipulating 3D Gaussians through control points and deformation MLPs. Another strategy involves constructing a 4D spatial-temporal representation. NeRFlow~\cite{du2021neural} learns a 4D spatial-temporal representation of a dynamic scene, while~\cite{xian2021space} represents a 4D space-time irradiance field that maps a spatial-temporal location to the color and volume density. HexPlane~\cite{cao2023hexplane}, K-Plane~\cite{DBLP:conf/cvpr/Fridovich-KeilM23} and 4D-GS~\cite{wu20244d} project 4D spatial-temporal space to multiple 2D planes.
\begin{figure*}[ht]
\centering
\includegraphics[width=2.1\columnwidth]{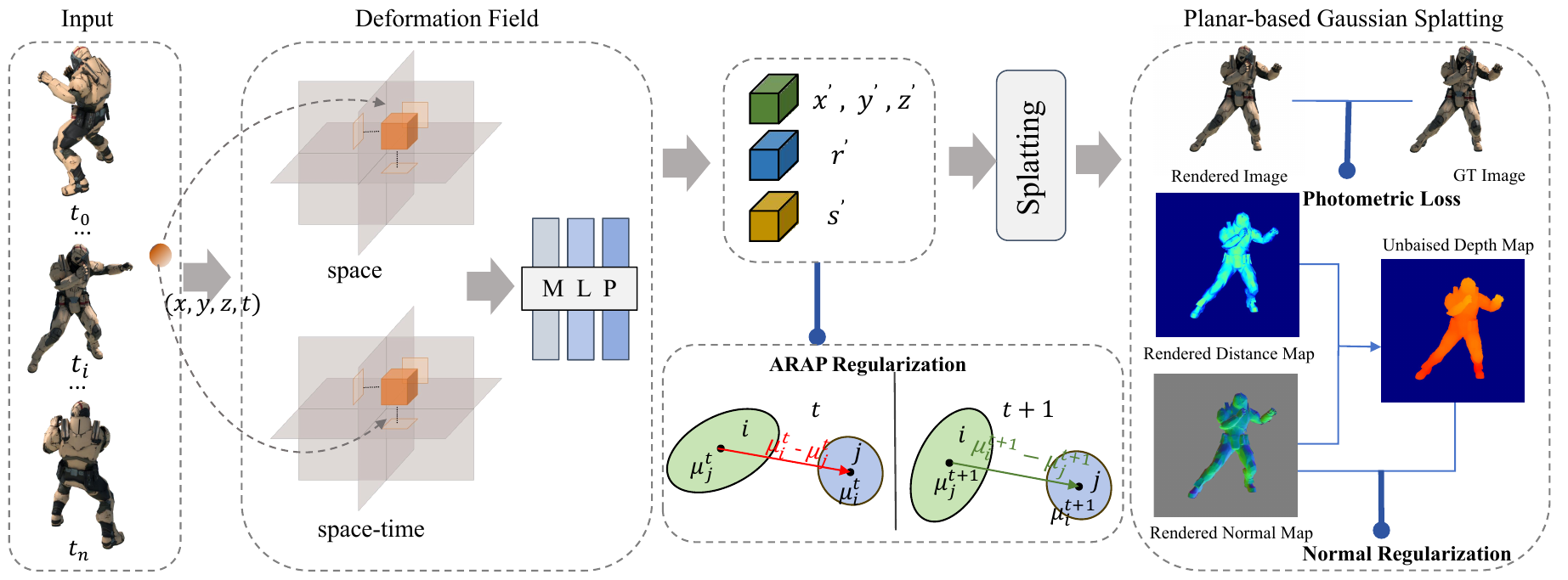}
\caption{\textbf{Overview of our method.} Firstly, in the deformation field, we represent the spatial and temporal information of dynamic objects in Hex-Plane and use an MLP to estimate the 3D Gaussian deformation. Subsequently, ARAP regularization is applied to ensure the local rigidity of the dynamic object at different moments. Finally, planar-based Gaussian splatting is used to obtain the unbiased depth map and render the transformed 3D Gaussian to images. }
\label{fig:framework}
 \vspace{-0.5cm}
\end{figure*}
However, these approaches primarily focus on the fidelity of rendered images to the ground truth, often overlooking the geometric reconstruction of dynamic objects.
\subsection{Surface Reconstruction} 
Surface Reconstruction~\cite{chen2024pgsr, Ye2024FedSurfGS, Ye2024DATAP-SfM, tang2024ndsdf, ye2022deflowslam, ye2023pvo, liu2021coxgraph, li2020saliency} focuses on generating accurate and detailed surface representations from sparse data. In recent years, the reconstruction of static scenes has been extensively studied. SuGaR~\cite{guedon2024sugar} introduces a regularization term that forces the Gaussians to be aligned with the scene surface. 2DGS~\cite{huang20242d} ensures view-consistent geometry modeling and rendering using 2D Gaussian primitives. PGSR~\cite{chen2024pgsr} introduces unbiased depth rendering to achieve high-quality surface reconstruction. However, these methods tend to fail when extended to dynamic scenes due to topology changes during deformation. To this end, 
Ub4D~\cite{johnson2023unbiased} models the topology of dynamic objects using implicit neural radiation fields.
More recently, several concurrent works have adapted 3D Gaussian for dynamic surface reconstruction. DG-Mesh~\cite{liu2024dynamic} encourages uniformly distributed Gaussian optimization by exploiting Gaussian mesh anchoring techniques. MaGS~\cite{ma2024reconstructing} captures the motion of each 3D Gaussian by employing a relative deformation field to model the relative displacement between the mesh and the 3D Gaussians. 
However, achieving smooth geometric surface reconstruction while maintaining high-quality rendered images remains challenging with these methods.
\begin{figure*}[ht]
\centering
\includegraphics[width=2.1\columnwidth]{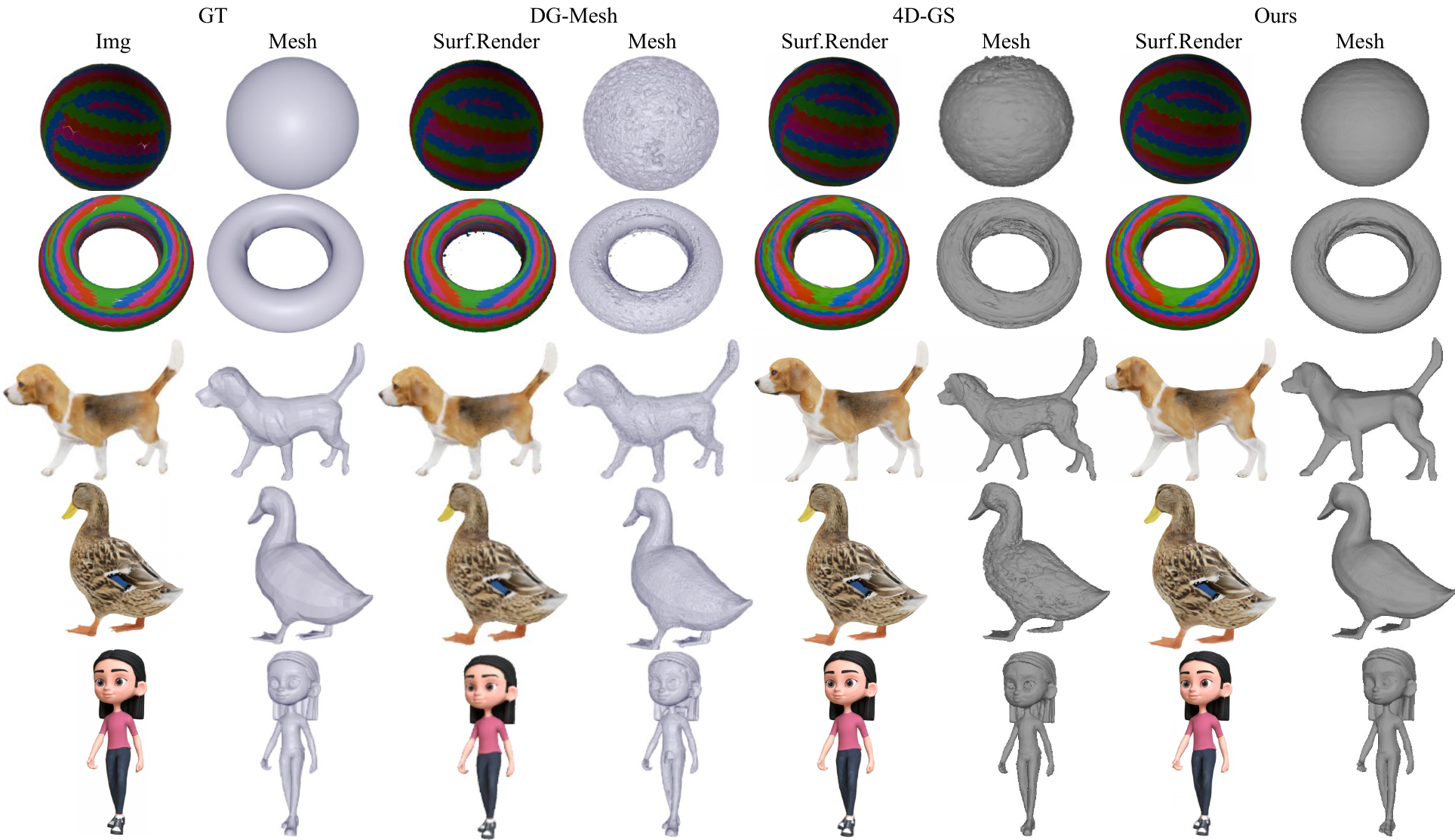}
\caption{\textbf{Qualitative comparison on the DG-Mesh dataset.} We present the visualizations of the reconstructed meshes and the rendered images. Our method demonstrates superior geometry and appearance compared to other baselines, which often display incomplete or noisy surfaces.}
\label{fig:dg-mesh}
 \vspace{-0.5cm}
\end{figure*}
\section{Preliminary}
3D Gaussians\cite{kerbl20233d} serve as a flexible and expressive representation of the scene. Each 3D Gaussian is characterized by an anisotropic covariance matrix $\boldsymbol{\Sigma}$ and the center $\mu$ of a point $p_i\in\mathcal{P}$, expressed as follows: 
\begin{equation}
{G}_i(x|\boldsymbol{\mu}_i,\boldsymbol{\Sigma}_{\boldsymbol{i}})=e^{-\frac12(\boldsymbol{x}-\boldsymbol{\mu}_i)^\top\boldsymbol{\Sigma}_i^{-1}(\boldsymbol{x}-\boldsymbol{\mu}_i)} ,
\end{equation}
To optimize the covariance matric $\boldsymbol{\Sigma}$, it can be decomposed into a scaling matrix $S_i \in\mathbb{R}^{3\times3}$ and a rotation matrix $R_{i}\in\mathbb{R}^{3\times3}$ as:
\begin{equation}
    \Sigma_i=R_iS_iS_i^\top R_i^\top,
    \label{3d_cov}
\end{equation}
where $R$ is a quaternion $r\in\mathbf{SO}(3)$, and S is a 3D vector s.
Moreover, 3D Gaussian represents color using spherical harmonic (SH) coefficients $sh$ and density through opacity $\sigma$. Thus, each gaussian ca be parameterized as $\mathcal{G}=\{G_{i}:\mu_{i},r_{i},s_{i},\sigma_{i},sh_{i}\}.$

When rendering a novel view, 3D Gaussian splatting enables fast $\alpha$-blendeing for rendering. As demonstrated in~\cite{zwicker2001surface}, the covariance matrix $\Sigma{^{\prime}}$ in the camera coordinates can be computed using the transformation matrix $W$ and the Jacobian matrix $J$ of the affine approximation of the projection transformation:
\begin{equation}
    \Sigma_i^{^{\prime}}=JW\Sigma_iW^{\top}J^{\top} ,
\end{equation}
For each pixel, the color $C \in\mathbb{R}^{3\times3}$ can be obtained via $\alpha$-blending:
\begin{equation}
    C=\sum_{i\in N}\alpha_ic_i\prod_{j=1}^{i-1}(1-\alpha_i),
\end{equation}
where $c_i$ is the SH color coefficients and $\alpha_i$ is the density calculated by:
\begin{equation}
    \alpha_i=\sigma_ie^{-\frac{1}{2}(x-\mu_i^{\prime})^T\Sigma_i^{\prime}(x-\mu_i^{\prime})} ,
\end{equation}
where $\mu_i^{\prime} = JW\mu$ is the center point in the camera coordinate. By optimizing the Gaussian parameters $\mathcal{G}=\{G_{i}: \mu_{i}, q_{i}, s_{i},\sigma_{i}, sh_{i}\}$ and adaptively adjusting the Gaussian density, real-time reconstruction of high-quality images and dynamic object surface reconstruction can be achieved.

\section{DynaSurfGS}
\subsection{Overview}
Given multiview RGB images of dynamic scenes, our goal is to achieve both high-quality dynamic surface reconstruction and superior rendering quality. 
As shown in Fig. \ref{fig:framework}, our framework incorporates deformation filed and planar-based Gaussian splatting given multiple images with corresponding camera poses and timestamps. Specifically, the deformation field comprises a Hex-Plane $E$ and a compact MLP decoder $D$. The Hex-Plane represents 4D volume, where the top three representing space and the bottom three capturing spatial and temporal variations. The 3D Gaussians are encoded by the Hex-Plane as $F = E(\mathcal{G})$, and then the MLP decoder to obtain the deformation of 3D Gaussians as $\triangle\mathcal{G} = D(F)$. The deformed 3D Gaussian is computed by adding these deformations to the original 3D Gaussians $\mathcal{G}$:
\begin{equation}
    (\mu^{'}, r^{'}, s^{'}, \sigma, sh) = (\mu + \triangle\mu, r+\triangle r, s+\triangle s, \sigma, sh).
\end{equation} 
Simultaneously, the Gaussian points across different time steps are constrained to adhere to the principle of local rigidity through ARAP regularization. For planar-based Gaussian splatting, we employ unbiased depth rendering to generate a planar distance map $L$ and a normal map $N$, which are subsequently converted into an unbiased depth map. The distance normal map is then computed using this unbiased depth map. The difference between the distance normal map and the normal map obtained through rasterization is minimized through normal regularization, which ensures smooth geometric surfaces for dynamic objects, as demonstrated in Fig. \ref{fig:vision_ablation} for "w/o ARAP Reg".
\begin{figure*}[ht]
\centering
\includegraphics[width=2\columnwidth]{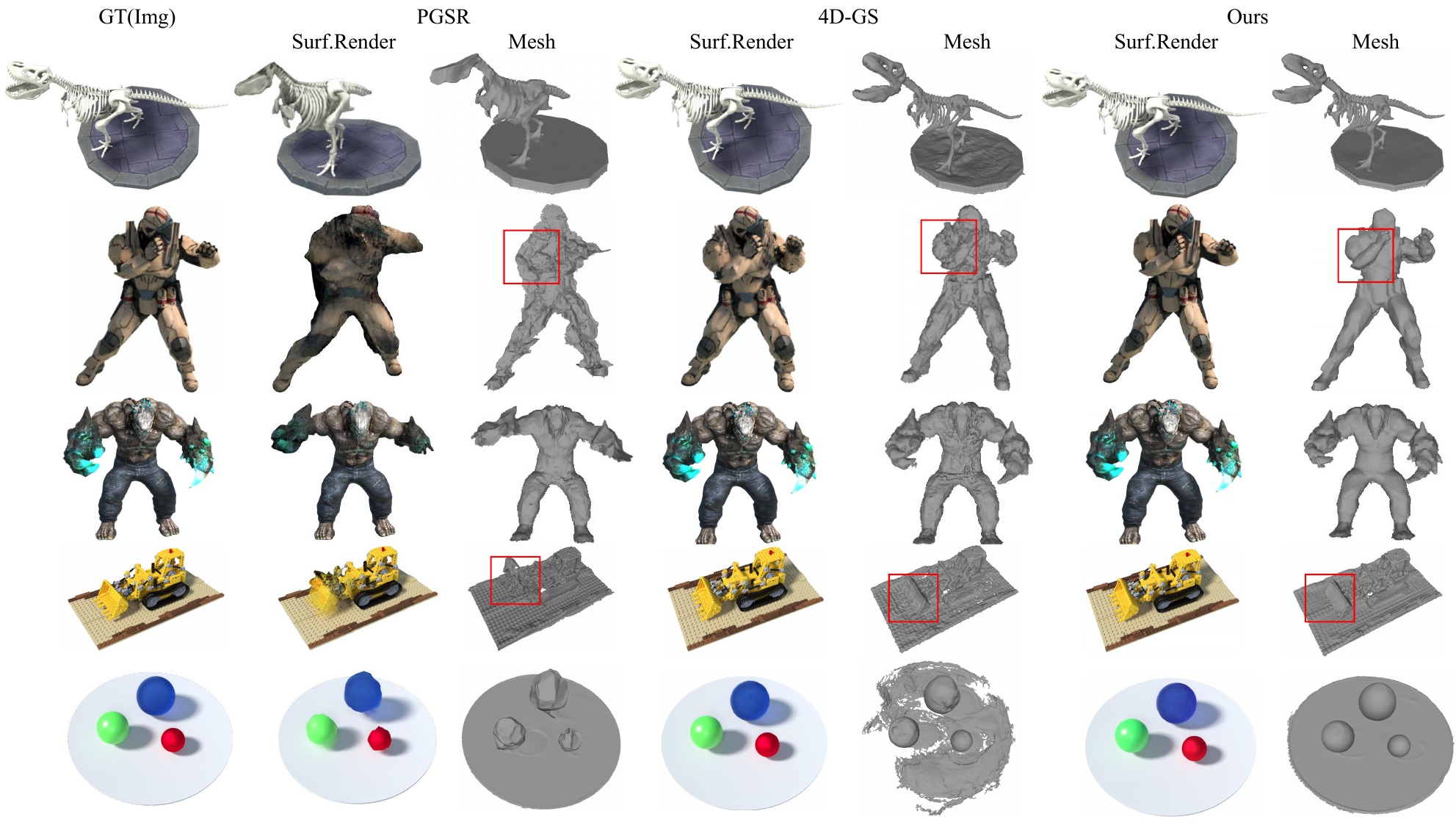}
\caption{\textbf{Qualitative comparison on the D-NeRF dataset.} We present the visualizations of the reconstructed meshes and the rendering images. Our method demonstrates superior geometry and appearance compared to other baselines, which often display incomplete or noisy surfaces.}
\label{fig:dnerf}
 \vspace{-0.5cm}
\end{figure*}
\subsection{Normal Regularization}
Inspired by PGSR~\cite{chen2024pgsr}, we leverage the unbiased depth rendering method based on 3DGS~\cite{kerbl20233d}. The normal map under the current viewpoint is rendered by $\alpha$-blending:
\begin{equation}
    N=\sum_{i\in N}T_iR_c^Tn_i\alpha_i,   T_i = \prod_{j=1}^{i-1}(1-\alpha_j),
    \label{rendered_normal}
\end{equation}
where $T_i$ is the cumulative opacity, $R_c$ denotes the rotation from the camera to the global world, and $n_i$ is the normal of the Gaussian. The distance from the camera center to the plane can be expressed as $l_i = (R_c^T(\mu_i-T_c))\boldsymbol{R}_c^T\boldsymbol{n}_i^T,$ where $\mu_i$ is the center of gaussian $G_i$ and $T_c$ is the camera center in the world. The distance map under the current viewpoint is rendered by $\alpha$-blending:
\begin{equation}
    \boldsymbol{L}=\sum_{i\in N}T_il_i\alpha_i, T_i = \prod_{j=1}^{i-1}(1-\alpha_j).
\end{equation}
Subsequently, the unbiased depth map is obtained by dividing the distance map by the normal map:
\begin{equation}
    L(m)=\frac L{N(m)\boldsymbol{K}^{-1}\tilde{\boldsymbol{m}}},
    \label{depth}
\end{equation}
where $m = [u,v]^T$ represents the 2D coordinates on the image plane, $\tilde{m}$ denotes the homogeneous coordinate of $m$, and $K$ is the intrinsic of camera.
Following~\cite{long2024adaptive, qi2020geonet++}, we constrain the local consistency of depth and normals based on the assumption of local planarity, where a pixel and its neighbors approximate a plane. Based on the depth map obtained from Eq. \ref{depth}, we sample four neighboring points using a fixed template. With these known depths, we compute the normal of the plane and repeat this process for the entire image to generate the normal from the depth map. The difference between this normal map and the rendered normal map as Eq. \ref{rendered_normal} is then minimized using $L1$ loss to ensure geometric consistency between local depths and normals. However, points in the edge region may not satisfy the planarity assumption, and to deal with this problem, we use image edges to approximate geometric edges. Specifically, for a pixel point $m$, we sample depth values from its four neighboring pixels (i.e., top, bottom, left, and right). These depth points are projected into the camera coordinate system to obtain the 3D points $\{\boldsymbol{M}_j|j=1,...,4\}$, and the local plane normal of pixel $m$ is computed as follows: 
\begin{equation}
    N_l(m)=\frac{(M_1-M_0)\times(M_3-M_2)}{|(M_1-M_0)\times(M_3-M_2)|},
\end{equation}
Finally, the normal regularization is calculated as follows:
\begin{equation}
    \mathcal{L}_{normal}=\frac{1}{W}\sum_{\boldsymbol{m}\in W}\left|\overline{\nabla\boldsymbol{I}}\right|^5\parallel N_l(\boldsymbol{m})-\boldsymbol{N}(\boldsymbol{m})\parallel_1,
\end{equation}
where $W$ denotes the set of image pixels and $N_m$ is defined by Eq. \ref{rendered_normal}. $\overline{\nabla\boldsymbol{I}}$ represents the image gradient normalized to a range of 0 to 1.
\subsection{ARAP Regularization}
To ensure the spatial consistency of a dynamic object across different time steps, we introduce the ARAP regularization $\mathcal{L}_{\mathrm{arap}}$, which guarantees local rigidity during motion. For each Gaussian point $i$, compute its set of $k$-nearest-neighbors ($k=10$). The influence of a neighboring Gaussian point $j$ on Gaussian point $i$ is represented by a weight $w_{ij}$, which decreases with increasing distance:
\begin{equation}
    w_{ij}=\frac{\tilde{w}_{ij}}{\sum\limits_{j\in K}\tilde{w}_{ij}},\text{where} \;  \tilde{w}_{ij}=\exp(-\frac{d_{ij}^2}{2o_j^2}),
\end{equation}
where $K={knn}_{i;k}$ ,\; $ d_{ij}$ denotes the distance between the center of the Gaussian $G_i$ and the neighboring Gaussian point $G_j$ and $o_j$ is the learned radius parameter of $G_j$.
To compute the ARAP regularization, we randomly sample two time steps $t_1$ and $t_2$. The deformation field is used to estimate the spatial positions $\mu_i^{t_1}, \mu_j^{t_1}, \mu_i^{t_2}, \mu_j^{t_2}$ of the Gaussian points $G_i$ and $G_j$ at $t_1$ and $t_2$, respectively. The local rigid rotation matrix $\tilde{R}i$ for Gaussian point $G_i$ is estimated as following:
\begin{equation}
    \tilde{R}_i=\underset{R\in\mathrm{SO}(3)}{\mathrm{arg}\operatorname*{min}}\sum_{j\in K}w_{ij}||(\mu_i^{t_1}-\mu_j^{t_1})-R(\mu_i^{t_2}-\mu_j^{t_2})||^2,
\end{equation}
where $R$ is a local frame rotation matrix as Eq. \ref{3d_cov}. The ARAP regularization, as illustrated in Fig .\ref{fig:framework}, is computed as:
\makeatletter
\renewcommand{\maketag@@@}[1]{\hbox{\m@th\normalsize\normalfont#1}}
\makeatother
\begin{small}
\begin{equation}
\hspace{-0.2cm}
    \mathcal{L}_{\mathrm{arap}}(\mu_i,t_1,t_2)=\sum_{j\in K}w_{ij}||(\mu_i^{t_1}-\mu_j^{t_1})-\tilde{R}_i(\mu_i^{t_2}-\mu_j^{t_2})||^2 .    
\end{equation}
\end{small}
\subsection{Optimization}
To optimize the deformation field, we utilize a combination of photometric loss, total-variational loss~\cite{sun2022direct, fang2022fast, cao2023hexplane} $\mathcal{L}_{tv}$, normal regularization, and ARAP regularization. As depicted in Fig. \ref{fig:framework}, the photometric loss is computed as the L1 loss between the rendered image and the ground truth.
\begin{equation}
    \mathcal{L}_{photo}= \parallel I - \tilde{I} \parallel_1 .
\end{equation}
The total training loss is a weighted sum of these loss terms:
\begin{equation}
    \mathcal{L}=\mathcal{L}_{photo}  + \mathcal{L}_{tv} + \lambda_{1}\mathcal{L}_{\mathrm{normal}}+ \lambda_{2}\mathcal{L}_{\mathrm{arap}}.
\end{equation}
Between 7000 and 9000 iterations, $\lambda_{1}$ and  $\lambda_{2}$ gradually increase, and after 9000 iterations, are set to 0.05 and 0.02, respectively.
\definecolor{lightorange}{RGB}{255, 223, 191}
\definecolor{lightyellow}{RGB}{255, 255, 191}
\begin{table}[t]
\vspace{-0.2cm}
\small
  \centering
  \caption{Metrics averaged over all scenes on the DG-Mesh dataset. ↑ means the higher, the better. \colorbox{pink}{Pink}, \colorbox{lightorange}{orange}, and \colorbox{lightyellow}{yellow} are used to indicate the best, second best, and third best, respectively. The rendering resolution is set to 800 x 800. $*$ represents the results we achieved by running the open source DG-Mesh ~\cite{liu2024dynamic}.}
    \resizebox{8cm}{!}{
  \begin{tabular}{cccc}
    \toprule
        Method & CD↓ & EMD↓ & PSNR↑ \\
\cmidrule(r){1-1} \cmidrule(r){2-4}
    D-NeRF   & 1.252 & 0.184  & 27.739\\
    K-Plane   & \colorbox{lightyellow}{1.068} & 0.147  & \colorbox{lightyellow}{31.142}\\
    HexPlane   & 1.954 & \colorbox{lightorange}{0.155}  & 30.103\\
    TiNeuVox-B  & 2.451 & 0.173  & \colorbox{lightorange}{31.428} \\
    DG-Mesh    & \colorbox{pink}{0.770}  & \colorbox{lightyellow}{0.128}  &  $15.774^*$   \\
\cmidrule(r){1-1} \cmidrule(r){2-4}
    Our & \colorbox{lightorange}{0.910} & \colorbox{pink}{0.123}  & \colorbox{pink}{32.508} \\
    \bottomrule
    \end{tabular}
    }
  \label{table: Quantitative on dgmesh}%
\end{table}%
\section{Experiments}
\subsection{Datasets}
To validate the effectiveness of our method, we conducted experiments on the D-NeRF~\cite{pumarola2021d} and DG-Mesh~\cite{liu2024dynamic} datasets and Ub4D~\cite{johnson2023unbiased} dataset. The D-NeRF dataset is a synthetic dataset comprising eight dynamic scenes, each scene has its corresponding precise camera parameter and timestamp information. These scenes consist of images with a resolution of 800×800 pixels, with each scene containing between 50 to 200 images for training. DG-Mesh~\cite{liu2024dynamic} dataset contains six dynamic scenes, each with 200 training and test images.  Ground truth meshes for each object are available, enabling quantitative and qualitative evaluations of the rendered meshes against the ground truth, thereby verifying the validity of our approach. For real-world data, we evaluate our method on Ub4D dataset. \\
\textbf{Evaluation Criterion.}
We evaluated the performance of dynamic view synthesis on three widely used image evaluation metrics, including the Peak Signal-to-Noise Ratio (PSNR), the Structural Similarity Index measure (SSIM), and the Learned Perceptual Image Patch Similarity (LPIPS)~\cite{zhang2018unreasonable}. To evaluate the quality of surface reconstruction of dynamic objects, we utilize the Chamfer Distance (CD) and the Earth Mover Distance (EMD) to measure the deviation between the reconstructed mesh and the ground truth. \\
\textbf{Implementation Details.}
Our model was trained on a single RTX 3090 GPU for 20k iterations and fine-tuned the optimization parameters according to the configuration in 3D-GS~\cite{kerbl20233d}. 
We use the TSDF Fusion algorithm~\cite{newcombe2011kinectfusion} to extract mesh and employ 4D-GS ~\cite{wu20244d} as our baseline. For more details, please refer to the supplementary material.
\begin{table}[tbp]
  \centering
  \tiny
  \caption{Metrics averaged over all scenes
  on the D-NeRF dataset.
 The rendering resolution is set to 800 x 800. The best results are bolded. Underline means second best.}
    \resizebox{8cm}{!}{
    \begin{tabular}{cccc}
    \toprule
       Method  & PSNR↑ & SSIM↑ & LPIPS↓\\
\cmidrule(r){1-1} \cmidrule(r){2-4}
D-NeRF & 30.43 & 0.9570 & 0.0704 \\
K-Plane & 30.67 & 0.9672 & 0.0453 \\
HexPlane & 31.02 & 0.9680 & 0.0392 \\
TiNeuVox & 31.35 & 0.9613 & 0.0519 \\
4D-GS & 34.06 & 0.9787 & \textbf{0.0218} \\
DG-Mesh & 23.00 & 0.9578 & 0.0522 \\
\cmidrule(r){1-1} \cmidrule(r){2-4}
    Ours & \textbf{34.31} & \textbf{0.9797} & \underline{0.0267}  \\  
    \bottomrule
    \end{tabular}
    }
  \label{table: dnerf}%
  \vspace{-0.3cm}
\end{table}%
\subsection{Qualitative Comparisons}
\begin{figure}[ht]
\centering
\includegraphics[width=\columnwidth]{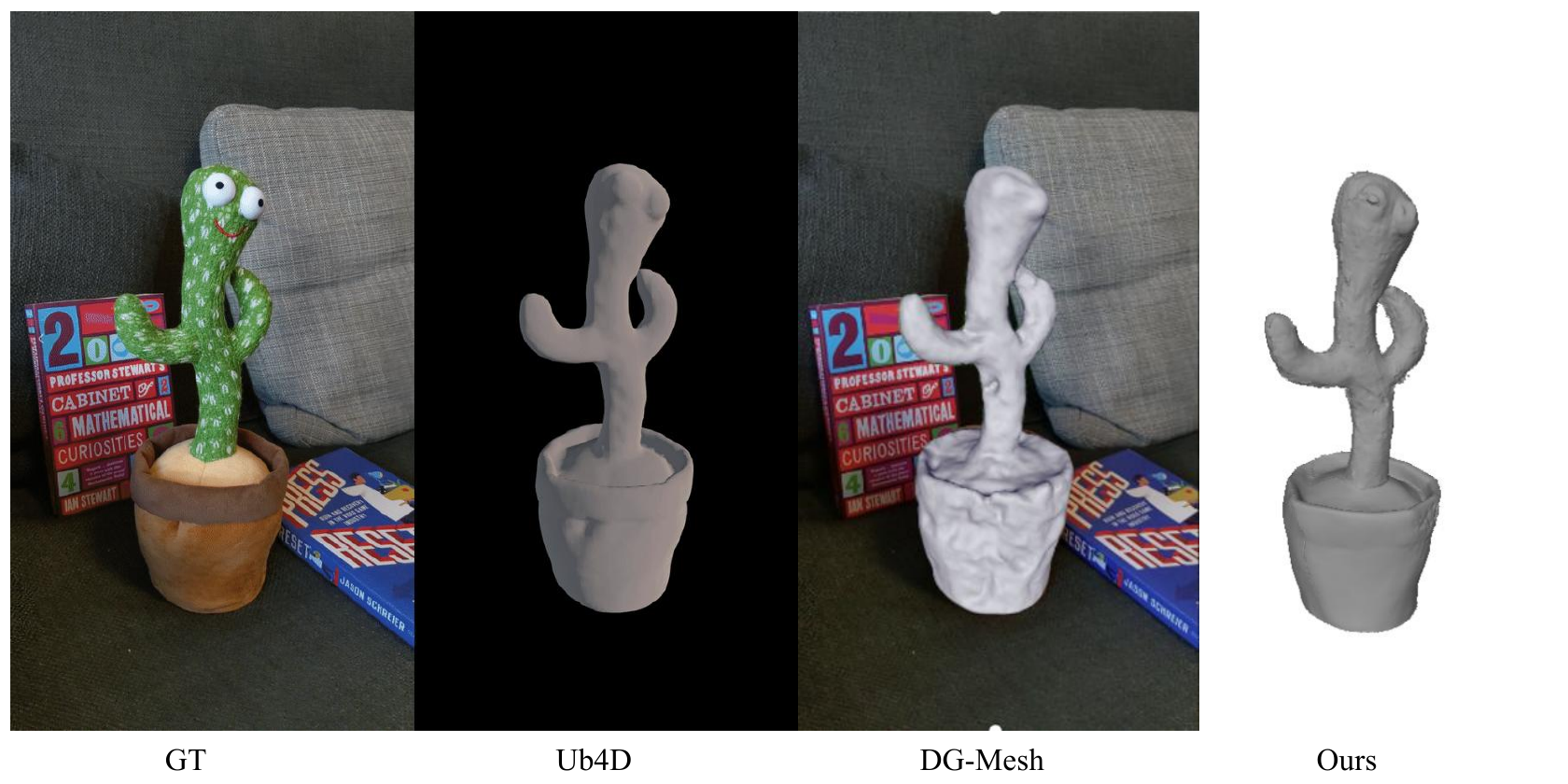}
\caption{\textbf{Qualitative comparison on the Ub4D dataset.} We provide the visualizations of the reconstructed meshes on the real data. Compared to other methods, our method is better at reconstructing geometric details such as eyes. }
\label{fig:unb4d}
\vspace{-0.5cm}
\end{figure}

To validate the effectiveness of our method, we conducted extensive qualitative comparison experiments on the D-NeRF~\cite{pumarola2021d}, DG-Mesh~\cite{liu2024dynamic}, and Ub4D ~\cite{johnson2023unbiased} datasets. For the D-NeRF dataset, we present visualizations of both the mesh and rendered images in Fig. \ref{fig:teaser}. While the RGB images rendered by 4D-GS~\cite{wu20244d} and MaGS~\cite{ma2024reconstructing} are visually comparable to our approach, the geometric surfaces of the reconstructed objects exhibit notable roughness. Our method outperforms DG-Mesh in both rendering and reconstruction quality, particularly evident in the second row of Fig. \ref{fig:dg-mesh}. As shown in Fig. \ref{fig:dnerf}, PGSR~\cite{chen2024pgsr}  excels in reconstructing static scene geometry, it fails to accurately reconstruct dynamic objects, such as the moving arm in the second row and the bucket in the fourth row (highlighted in red boxes). This shortcoming arises because PGSR considers only the spatial geometry of 3D objects, neglecting the temporal information critical for dynamic scenes. Moreover, the geometric surfaces reconstructed by 4D-GS are considerably rough, particularly in the bucket, as the method prioritizes rendering quality over geometric accuracy. In contrast, our approach delivers superior results in both dynamic surface reconstruction and image rendering. Fig. \ref{fig:dg-mesh} shows the visualization comparison of our method with 4D-GS and the concurrent work DG-Mesh on the DG-Mesh dataset. Our method produces more accurate and smoother geometric surfaces, a result attributable to the implementation of normal regularization. The surfaces generated by our method are significantly closer to the ground truth mesh compared to the rougher surfaces produced by 4D-GS and DG-Mesh, particularly evident in the first and second rows. For real data, our method has better geometric detail than other baselines, especially in the dolls' eyes, as shown in Fig.\ref{fig:unb4d}.

\begin{table}[tbp]
  \centering
  \tiny
  \caption{Average metrics for all scene ablation study on the DG-Mesh dataset,
  with a rendering resolution set to 800 x 800, our method achieves high-precise geometry while maintaining comparable high-quality images.}
    \resizebox{8cm}{!}{
    \begin{tabular}{cccc}
    \toprule
       Method & CD↓ & EMD↓ & PSNR↑\\
\cmidrule(r){1-1} \cmidrule(r){2-4}
    baseline   & {0.618} & {0.114}  & {41.53}   \\
    w/o ARAP Reg   & {0.617} & {0.111}  & {40.57}  \\
    w/o Normal Reg   & {0.617} & {0.114}  & \textbf{41.78} \\
\cmidrule(r){1-1} \cmidrule(r){2-4}
    Ours & \textbf{0.609} & \textbf{0.110}  & {40.74}  \\  
    \bottomrule
    \end{tabular}
    }
  \label{table: ablations 2}%
  \vspace{-0.2cm}
\end{table}%
\begin{table}[tbp]
  \centering
  \small
  \caption{Average metrics for all scene ablation study on the D-NeRF dataset. The rendering resolution is 800 x 800.}
    \resizebox{\linewidth}{!}{
    \begin{tabular}{cccc}
    \toprule
       Method  & PSNR↑ & SSIM↑ & LPIPS↓\\
\cmidrule(r){1-1} \cmidrule(r){2-4}
Baseline & 34.06 & 0.9787 & \textbf{0.0218} \\
w./ Normal Reg & 33.93 & 0.9786 & 0.0268 \\
w./ Normal Reg + ARAP Reg & \textbf{34.31} & \textbf{0.9797} & 0.0267 \\
    \bottomrule
    \end{tabular}
    }
  \label{table: ablations on the d-nerf}%
\end{table}%
\subsection{Quantitative Comparisons}
In Tab. \ref{table: Quantitative on dgmesh}, we compare our method with several state-of-the-art methods, including D-NeRF~\cite{pumarola2021d}, K-plane~\cite{DBLP:conf/cvpr/Fridovich-KeilM23}, HexPlane~\cite{cao2023hexplane}, TiNeuVox-B~\cite{fang2022fast}, DG-Mesh~\cite{liu2024dynamic} on the DG-Mesh dataset . As shown in Tab. \ref{table: Quantitative on dgmesh}, we calculated the average of the metrics CD and EMD for evaluating the mesh quality of all dynamic object surface reconstruction and the average of the PSNR for evaluating the quality of the rendered RGB images in the DG-Mesh dataset. For per-scene results of DG-Mesh dataset, please refer to the supplementary material. In Tab. \ref{table: Quantitative on dgmesh}, our method generally achieves CD and EMD values comparable to DGmesh, while our rendered images achieve optimal PSNR values, indicating that our method is capable of achieving high-quality rendered images as well as high-fidelity reconstruction of dynamic object surfaces. Fig. \ref{fig:dnerf} also shows that our method achieves accurate dynamic surface reconstruction on the D-NeRF dataset.
\begin{figure}[t]
\centering
\includegraphics[width=\columnwidth]{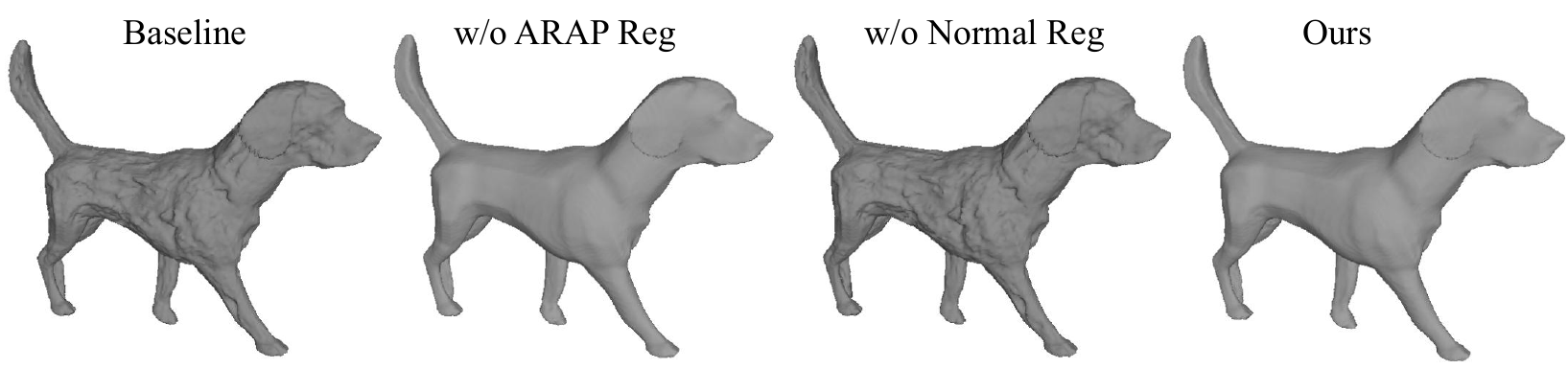}
\caption{\textbf{Ablation study on the DG-Mesh dataset.} Using only the ARAP regularization results in a rough surface for the reconstructed geometry, whereas relying solely on normal regularization produces a smoother geometric surface but compromises the quality of the rendered 2D images, as evidenced in Tab. \ref{table: ablations 2}. The integration of both ARAP and normal regularization produces high-fidelity dynamic surface reconstruction and photorealistic rendered images.}
\label{fig:vision_ablation}
\end{figure}
\subsection{Ablation Studies}
\textbf{Normal Regularization} Normal regularization significantly enhances the smoothness of geometric surfaces and provides a robust geometric initialization. As illustrated in Fig. \ref{fig:vision_ablation},  normal regularization is essential for accurate geometric reconstruction, as seen in the "w/o ARAP Reg" scenario. However, it is observed that normal regularization slightly reduces the PSNR of the rendered images compared to the baseline, as shown in Tab. \ref{table: ablations 2}. To address this issue while maintaining rigid body consistency across different moments, we introduce ARAP regularization.\\
\textbf{ARAP Regularization} The experiments conducted on the D-NeRF~\cite{pumarola2021d} dataset, presented in Tab. \ref{table: ablations on the d-nerf}, shows that the results for "$w./$ Normal Reg" degrade in performance when ARAP regularization is excluded, which validates the effectiveness of the ARAP constraint. In Tab. \ref{table: ablations 2}, the results for "w/o Normal Reg" indicate that while the best PSNR value is achieved using only ARAP regularization, the CD and EMD values are poor, indicating that although the rendered images are high-quality, the reconstructed geometries are suboptimal, as further confirmed in Fig. \ref{fig:vision_ablation} for "w/o Normal Reg". Therefore, the combination of ARAP and normal regularization is crucial for producing photorealistic images while reconstructing high-fidelity geometry.\\
\section{Limitations and Future Work} 
 While our method achieves highly accurate geometric reconstruction and produces photorealistic images, several challenges persist.
 Firstly, our regularization often requires a trade-off between image quality and geometry accuracy, which can sometimes lead to over-smoothing in certain areas. Future research could explore adaptive regularization methods that dynamically adjust parameters based on local image features, aiming to prevent excessive smoothing while maintaining image quality. Secondly, areas with missing or limited views cannot be fully reconstructed geometrically, resulting in incomplete or less accurate geometry. To address this, integrating large-scale pre-trained diffusion models to generate geometries from unseen viewpoints could be a promising direction for further investigation.
\section{Conclusion}
    In this work, we present DynaSurfGS, a novel method for reconstructing high-accuracy geometry and rendering high-quality images of dynamic objects from a monocular video. Our approach leverages planar-based Gaussian splatting to obtain unbiased depth maps, which are subsequently used to constrain the smooth reconstruction of geometric surfaces for dynamic objects through normal regularization. Additionally, we incorporate ARAP regularization to maintain spatial consistency by constraining rigid body motion across different moments. We validated the effectiveness of our method on the D-NeRF, DG-Mesh, and Ub4D datasets with extensive experiments that demonstrated its superior performance compared to existing baselines. \\
   
\bibliography{aaai25}

\newpage
\clearpage

\appendix

\section{Oultine}
In the supplementary file, we provide more implementation details and more results not elaborated in our paper due to this paper length limit:
 \begin{itemize}
\item[$\bullet$] Sec.~\cww{S1}: per-scene qualitative experiments. 
\item[$\bullet$] Sec.~\cww{S2}: per-scene quantitative experiments.
\item[$\bullet$] Sec.~\cww{S3}: more implementation details.
\end{itemize}

\section{S1. Per-scene qualitative experimennts}
We produced a 360-degree video showcasing the rendered images and meshes of the dynamic objects reconstructed by our approach, along with comparative results from DG-Mesh~\cite{liu2024dynamic} and 4D-GS~\cite{wu20244d}. Fig. \cww{S2.1} and Fig. \cww{S2.2} present screenshots from the video on the D-NeRF~\cite{pumarola2021d} and DG-Mesh~\cite{liu2024dynamic} datasets, respectively, with the full video available in the video file. As shown in Fig. \cww{S2.1}, the D-NeRF~\cite{pumarola2021d} dataset lacks ground truth meshes, so an input image is used as the ground truth. DG-Mesh reconstructs the surface of a dynamic object that is not smooth enough, and the rendered image is blurred. While 4D-GS renders high-quality RGB images, the reconstructed geometric surfaces are rough and even incomplete. In contrast, our method demonstrates superior performance by producing high-quality rendered images and precise dynamic surface reconstructions. Fig. \cww{S2.2} further demonstrates that our method achieves both high-quality image rendering and dynamic surface reconstruction compared to DG-Mesh and 4D-GS.

\section{S2. Per-scene quantitative experiments}
Quantitative experiments on the averages of the D-NeRF ~\cite{pumarola2021d} and DG-Mesh~\cite{liu2024dynamic} datasets have been presented in the main text, and the results of the quantitative experiments on a scene-by-scene are presented here. Tab. \ref{table: Quantitative dgmesh} illustrates the quantitative results of our method compared to existing methods on the DG-Mesh~\cite{liu2024dynamic} dataset. $*$ denotes the results obtained by running the open-source DG-Mesh~\cite{liu2024dynamic} code. As shown in Tab. \ref{table: Quantitative dgmesh}, our method outperforms DG-Mesh~\cite{liu2024dynamic} across most quantitative metrics for dynamic surface reconstruction. In particular, for the rendering image quality method, the average of our method is $2$ times better than DG-Mesh. Since the D-NeRF dataset does not have the ground truth meshes, we quantitatively evaluate the rendered image in the D-NeRF dataset. Tab. \ref{table: dnerf} demonstrates that the image quality rendered by our method generally surpasses that of 4D-GS~\cite{wu20244d}. Moreover, the dynamic meshes produced by our approach are significantly smoother than those generated by 4D-GS, as depicted in Fig. \cww{S2.1}. In addition, we conducted ablation experiments on the D-NeRF dataset for all objects, as shown in Tab. \ref{table: ablation}.
\begin{figure*}[htbp]
\centering
\includegraphics[width=2\columnwidth]{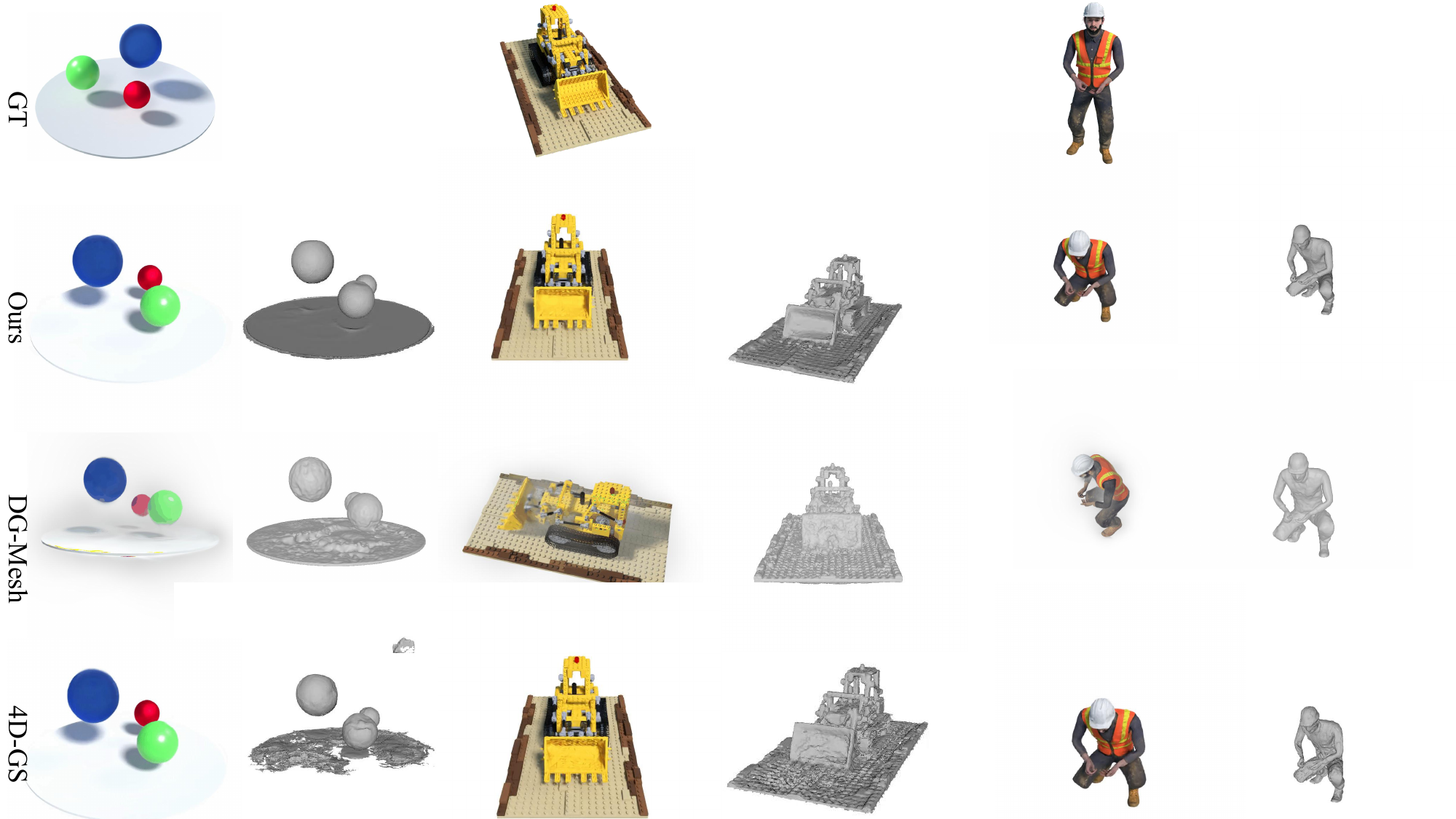}
 \vspace{-0.5cm}
\end{figure*}
\begin{figure*}[htp]
\centering
\includegraphics[width=2\columnwidth]{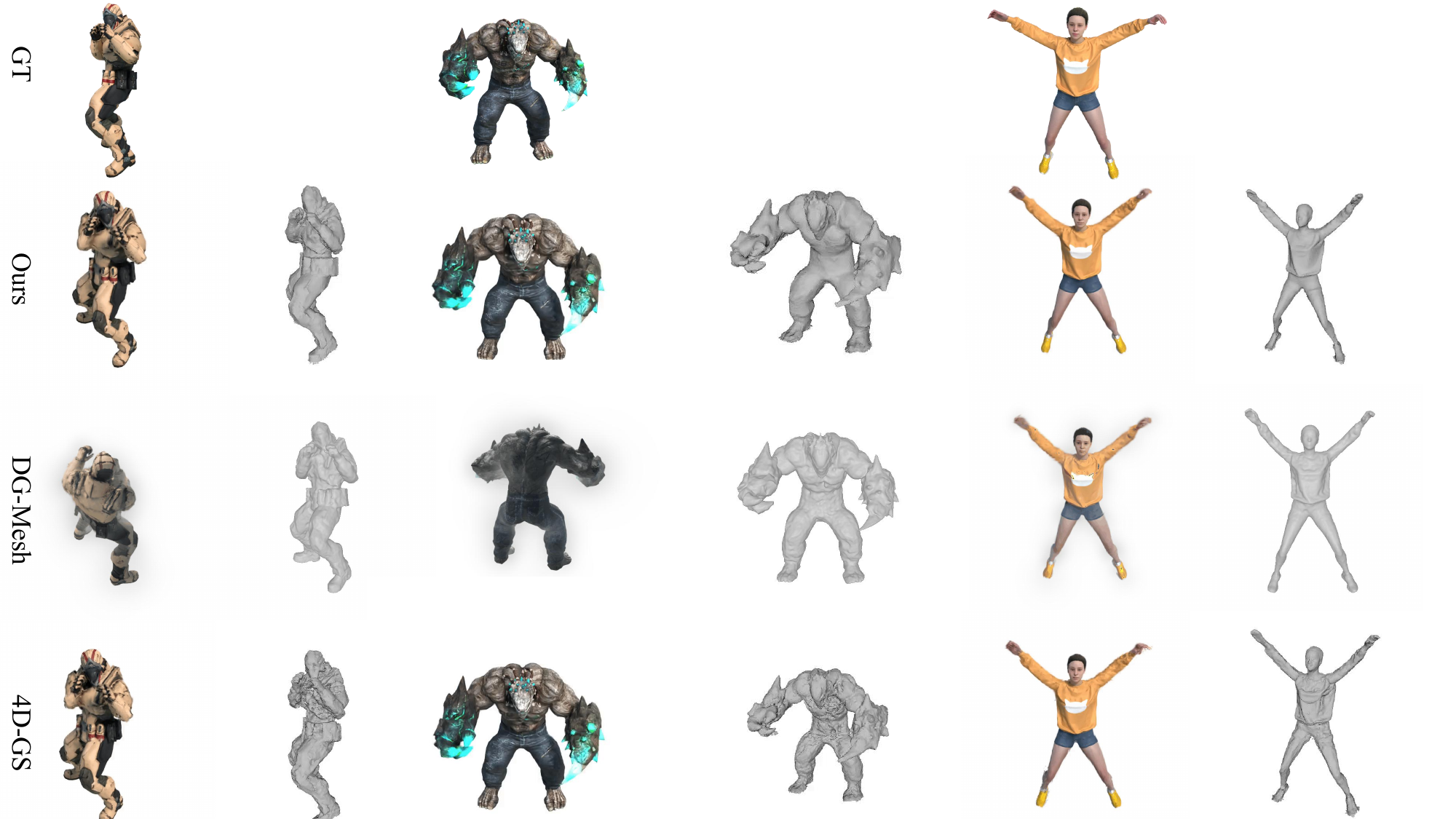}
 \vspace{-0.5cm}
\end{figure*}

\begin{figure*}[htbp]
\centering
\includegraphics[width=1.8\columnwidth]{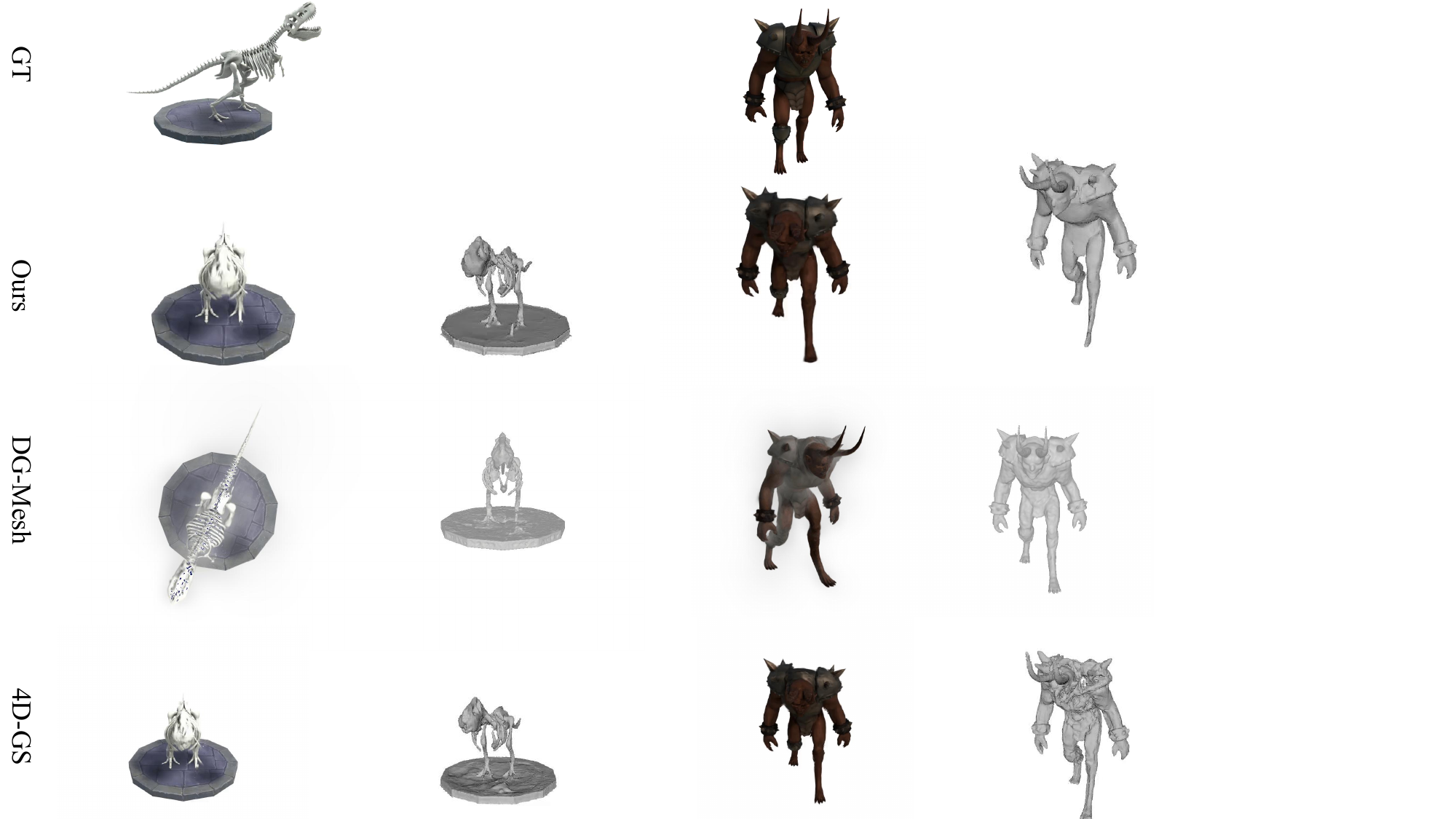} \\
Figure \cww{S2.1}: Per-scene qualitative experiments on the D-NeRF~\cite{pumarola2021d} dataset.
\label{fig:supp_1}
\end{figure*}


\begin{figure*}[htp]
\centering
\includegraphics[width=2\columnwidth]{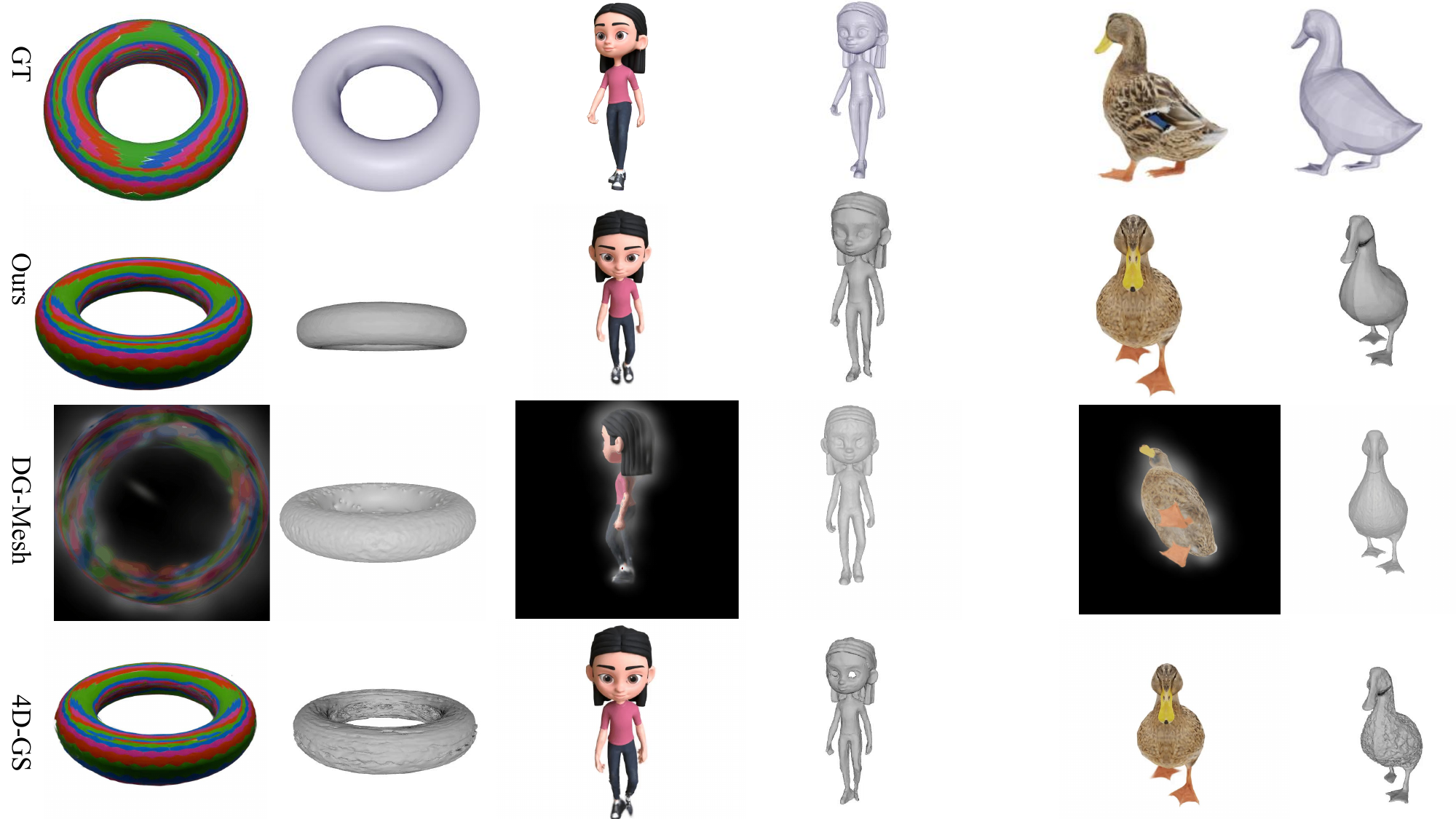}
\end{figure*}

\begin{figure*}[tbp]
\centering
\includegraphics[width=2\columnwidth]{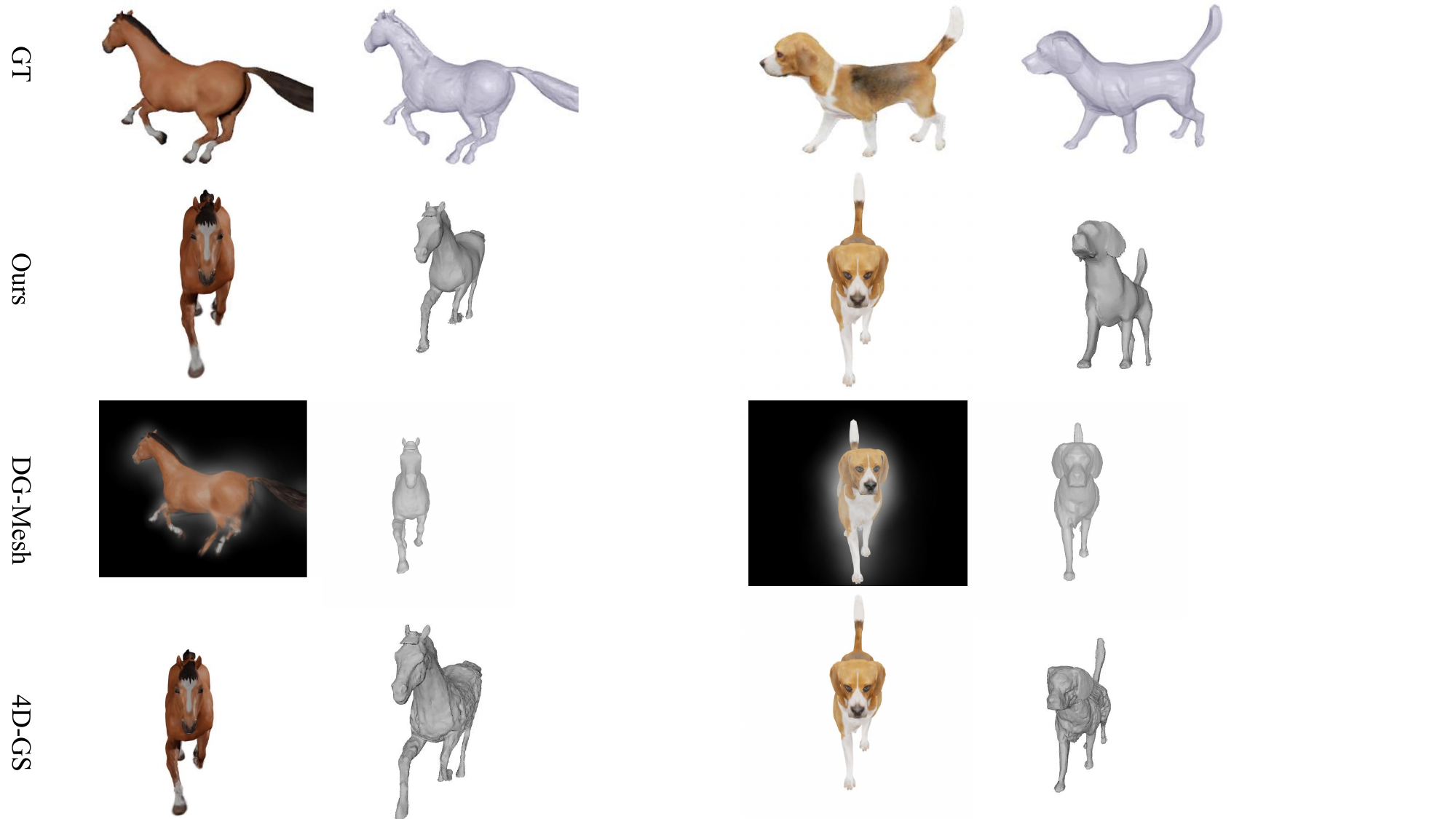} \\
Figure \cww{S2.2}: Per-scene qualitative experiments on the DG-Mesh~\cite{liu2024dynamic} dataset.
\label{fig:supp_5}
\end{figure*}

\definecolor{lightorange}{RGB}{255, 223, 191}
\definecolor{lightyellow}{RGB}{255, 255, 191}
\begin{table*}[tbp]
\Huge
\renewcommand{\thetable}{\cww{S1.1}}
  \centering
  \caption{The qualitative comparison of our method with existing methods on the DG-Mesh~\cite{liu2024dynamic} dataset, ↑ means the higher, the better. \colorbox{pink}{Pink}, \colorbox{lightorange}{orange}, and \colorbox{lightyellow}{yellow} are used to indicate the best, second best, and third best, respectively. The rendering resolution is set to 800 x 800. $*$ represents the results we achieved by running the open source DG-Mesh code ~\cite{liu2024dynamic}.}
    \resizebox{\linewidth}{!}{
    \begin{tabular}{cccccccccc}
    \toprule
    \multicolumn{1}{c}{\multirow{2}[2]{*}{Method}} &
    \multicolumn{3}{c}{Beagle} & 
    \multicolumn{3}{c}{Girlwalk} & 
    \multicolumn{3}{c}{Duck}  \\
    \cmidrule(r){2-4} \cmidrule(r){5-7} 
    \cmidrule(r){8-10} 
      &  CD↓ & EMD↓ & PSNR↑ & CD↓ & EMD↓ & PSNR↑ & CD↓ & EMD↓& PSNR↑ \\
\cmidrule(r){1-1} \cmidrule(r){2-4}
\cmidrule(r){5-7} \cmidrule(r){8-10}
    D-NeRF   & 1.001 & 0.149  & 34.470  & 0.601 & 0.190 & 28.632 & \colorbox{lightyellow}{0.934} & 0.073 & 29.785 \\
    K-Plane   & \colorbox{lightyellow}{0.810} & 0.122  & \colorbox{lightyellow}{38.329} & \colorbox{lightorange}{0.495} & 0.173 & \colorbox{lightyellow}{32.116} & 1.085 & \colorbox{lightyellow}{0.055} & \colorbox{lightyellow}{33.360}   \\
    HexPlane   & 0.870 & \colorbox{lightorange}{0.115}  & 38.034 & 0.597 & \colorbox{lightyellow}{0.155} & 31.771 & 2.161 & 0.090 & 32.108\\
    TiNeuVox-B  & 0.874 & 0.129  & \colorbox{lightorange}{38.972} & \colorbox{lightyellow}{0.568} & 0.184 & \colorbox{lightorange}{32.806} & 0.969 & 0.059 & \colorbox{lightorange}{34.326} \\
    DG-Mesh    & \colorbox{lightorange}{0.639}  & \colorbox{lightyellow}{0.117}  &  $16.135^*$     & 0.726 & \colorbox{lightorange}{0.136} & $17.525^*$     & \colorbox{pink}{0.790} & \colorbox{lightorange}{0.047} & $16.511^*$  \\
\cmidrule(r){1-1} \cmidrule(r){2-4}
\cmidrule(r){5-7} \cmidrule(r){8-10}
    \textbf{Our} & \colorbox{pink}{0.609} & \colorbox{pink}{0.110}  & \colorbox{pink}{40.744} & \colorbox{pink}{0.443} & \colorbox{pink}{0.128} & \colorbox{pink}{33.308} & \colorbox{lightorange}{0.806} & \colorbox{pink}{0.047} & \colorbox{pink}{36.310}  \\
      \bottomrule
      \multicolumn{1}{c}{\multirow{2}[2]{*}{Method}} &
    \multicolumn{3}{c}{Horse} & 
    \multicolumn{3}{c}{Bird} & 
    \multicolumn{3}{c}{Torus2sphere}  \\
    \cmidrule(r){2-4} \cmidrule(r){5-7} 
    \cmidrule(r){8-10} 
      &  CD↓ & EMD↓ & PSNR↑ & CD↓ & EMD↓ & PSNR↑ & CD↓ & EMD↓& PSNR↑ \\
\cmidrule(r){1-1} \cmidrule(r){2-4}
\cmidrule(r){5-7} \cmidrule(r){8-10}
    D-NeRF   & 1.685 & 0.280  & 25.474  & \colorbox{lightyellow}{1.532} & 0.163 & 23.848 & \colorbox{lightyellow}{1.760} & 0.250 & 24.227 \\
    K-Plane   & \colorbox{lightyellow}{1.480} & 0.239  & \colorbox{lightyellow}{28.111} & \colorbox{lightorange}{0.742} & \colorbox{lightorange}{0.131} & \colorbox{lightyellow}{23.722} & 1.793 & \colorbox{pink}{0.161} & \colorbox{pink}{31.215}   \\
    HexPlane   & 1.750 & \colorbox{lightorange}{0.199}  & 26.799 & 4.158 & 0.178 & 22.189 & 2.190 & 0.190 & \colorbox{lightorange}{29.714}\\
    TiNeuVox-B  & 1.918 & 0.246  & \colorbox{lightorange}{28.161} & 8.264 & 0.215 & \colorbox{lightorange}{25.546} &2.115 & 0.203 & 28.756 \\
    DG-Mesh    & \colorbox{lightorange}{0.299}  & \colorbox{lightyellow}{0.168}  &  $15.484^*$     & \colorbox{pink}{0.557} & \colorbox{pink}{0.128} &  $17.151^*$    & \colorbox{pink}{1.607} & \colorbox{lightyellow}{0.172} & $11.835^*$  \\
\cmidrule(r){1-1} \cmidrule(r){2-4}
\cmidrule(r){5-7} \cmidrule(r){8-10}
    \textbf{Our} & \colorbox{pink}{0.296} & \colorbox{pink}{0.145}  & \colorbox{pink}{28.680} & 1.631 & \colorbox{lightyellow}{0.138} & \colorbox{pink}{26.876} & \colorbox{lightorange}{1.675} & \colorbox{lightorange}{0.171} & \colorbox{lightyellow}{29.131}  \\
    \bottomrule
    \end{tabular}
    }
  \label{table: Quantitative dgmesh}%
\end{table*}%

\begin{table*}[t]
\Huge
\renewcommand{\thetable}{\cww{S1.2}}
\vspace{-5cm}
  \centering
  \caption{Per-scene quantitative experiments on the D-NeRF~\cite{pumarola2021d} dataset.}
    \resizebox{\linewidth}{!}{
    \begin{tabular}{ccccccccccccc}
    \toprule
    \multicolumn{1}{c}{\multirow{2}[2]{*}{Method}} & \multicolumn{3}{c}{Bouncingballs} & \multicolumn{3}{c}{Hook} & \multicolumn{3}{c}{Hellwarrior} & \multicolumn{3}{c}{Jumpingjacks}\\
    \cmidrule(r){2-4} \cmidrule(r){5-7} 
    \cmidrule(r){8-10} \cmidrule(r){11-13} 
        & PSNR↑ & SSIM↑ & LPIPS↓ & PSNR↑ & SSIM↑ & LPIPS↓ & PSNR↑ & SSIM↑ & LPIPS↓ & PSNR↑ & SSIM↑ & LPIPS↓ \\
        \cmidrule(r){1-1} \cmidrule(r){2-4}
\cmidrule(r){5-7} \cmidrule(r){8-10}
\cmidrule(r){11-13}
    D-NeRF   & {38.93} & {0.987} & {0.1074}  & {29.25} & {0.968} & {0.1120} & {25.02} & {0.955} & {0.0633} & {32.80} & {0.981} & {0.0381} \\
    K-Plane  &40.05&0.9934&0.0322&28.12&0.9489&0.0662&24.58&0.9520&0.0824&31.11&0.9708&0.0468 \\
     HexPlane &39.86&0.9915&0.0323&28.63&0.9572&0.0505&24.55&0.9443&0.0732&31.31&0.9729&0.0398 \\
    TiNeuVox &40.23&0.9926&0.0416&28.63&0.9433&0.0636&27.10&0.9638&0.0768&33.49&0.9771&0.0408 \\
   4D-GS   & 40.62 & 0.9942  & 0.0155 & 32.73 & 0.9760 & \textbf{0.0272} & 28.71 & 0.9733 & 0.0369 & 35.42 & 0.9857 & \textbf{0.0128} \\
    DG-Mesh & 23.15 & 0.9727  & 0.0658 & 22.12 & 0.9486 & 0.0555 & 19.57 & 0.9227 & 0.0732 & 26.29 & 0.9710 & 0.0458 \\
\cmidrule(r){1-1} \cmidrule(r){2-4}
\cmidrule(r){5-7} \cmidrule(r){8-10}
\cmidrule(r){11-13}
     Ours & \textbf{40.92} & \textbf{0.9948}  & \textbf{0.0139} & \textbf{32.97} & \textbf{0.9773} & 0.0277 & \textbf{29.45} & \textbf{0.9758} & \textbf{0.0360} & \textbf{35.49} & \textbf{0.9864} & 0.0202 \\

    \bottomrule
    \multicolumn{1}{c}{\multirow{2}[2]{*}{Method}} &  \multicolumn{3}{c}{Mutant}  & \multicolumn{3}{c}{Standup} & \multicolumn{3}{c}{Trex} & \multicolumn{3}{c}{Lego}  \\
    \cmidrule(r){2-4} \cmidrule(r){5-7} 
    \cmidrule(r){8-10} \cmidrule(r){11-13}
        & PSNR↑ & SSIM↑ & LPIPS↓ & PSNR↑ & SSIM↑ & LPIPS↓ & PSNR↑ & SSIM↑ & LPIPS↓ & PSNR↑ & SSIM↑ & LPIPS↓  \\
\cmidrule(r){1-1} \cmidrule(r){2-4}
\cmidrule(r){5-7} \cmidrule(r){8-10}
\cmidrule(r){11-13}
D-NeRF    & {31.29} & {0.978} & {0.0212}& {32.79} & {0.983} & {0.0241} & {31.75} & {0.974} & {0.0367} & 21.64 & 0.83 & 0.16 \\
    K-Plane   &32.50&0.9713&0.0362&33.10&0.9793&0.0310&30.43&0.9737&0.0343&\textbf{25.49}&\textbf{0.9483}&\textbf{0.0331} \\
     HexPlane &33.67&0.9802&0.0261&34.40&0.9839&0.0204&30.67&0.9749&0.0273&25.10&0.9388&0.0437 \\
    TiNeuVox &30.87&0.9607&0.0474&34.61&0.9797&0.0326&31.25&0.9666&0.0478&24.65&0.9063&0.0648 \\
    4D-GS &  37.59 & 0.9880 & \textbf{0.0167} & 38.11 & 0.9898  & \textbf{0.0138}  & \textbf{34.23} & \textbf{0.9850} & \textbf{0.0131} & 25.03 & 0.9376 & 0.0382 \\
    DG-Mesh  & 22.45 & 0.9609 & 0.0391& 24.12 & 0.9676  & 0.0350 & 23.29 & 0.9611 & 0.0512  & 22.32 & 0.0908 & 0.0565 \\
\cmidrule(r){1-1} \cmidrule(r){2-4}
\cmidrule(r){5-7} \cmidrule(r){8-10}
\cmidrule(r){11-13}
     Ours & \textbf{38.61} & \textbf{0.9903} & 0.0150 & \textbf{37.76}  & \textbf{0.9884} & 0.0191 & 34.21 & 0.9848 & 0.0230 & 25.02 & 0.9374 & 0.0583  \\
    \bottomrule
    \end{tabular}}
  \label{table: dnerf}%
      
\end{table*}%

\begin{table*}[tbp]
\Huge
\renewcommand{\thetable}{\cww{S1.3}}
\vspace{-13cm}
  \centering
  \caption{Per-scene ablation studies on the D-NeRF dataset}
    \resizebox{\linewidth}{!}{
    \begin{tabular}{ccccccccccccc}
    \toprule
    \multicolumn{1}{c}{\multirow{2}[2]{*}{Method}} & \multicolumn{3}{c}{Bouncingballs} & \multicolumn{3}{c}{Hook} & \multicolumn{3}{c}{Hellwarrior} & \multicolumn{3}{c}{Jumpingjacks}\\
    \cmidrule(r){2-4} \cmidrule(r){5-7} 
    \cmidrule(r){8-10} \cmidrule(r){11-13} 
        & PSNR↑ & SSIM↑ & LPIPS↓ & PSNR↑ & SSIM↑ & LPIPS↓ & PSNR↑ & SSIM↑ & LPIPS↓ & PSNR↑ & SSIM↑ & LPIPS↓ \\
        \cmidrule(r){1-1} \cmidrule(r){2-4}
\cmidrule(r){5-7} \cmidrule(r){8-10}
\cmidrule(r){11-13}
    Baseline   & 40.62 & 0.9942  & 0.0155 & 32.73 & 0.9760 & \textbf{0.0272} & 28.71 & 0.9733 & 0.0369 & 35.42 & 0.9857 & \textbf{0.0128} \\
\cmidrule(r){1-1} \cmidrule(r){2-4}
\cmidrule(r){5-7} \cmidrule(r){8-10}
\cmidrule(r){11-13}
    w./ Normal Reg   & 40.05 & 0.9945  & 0.0142 & 32.71 & 0.9765 & 0.0282 & 29.14 & 0.9749 & 0.0365 & 35.24 & 0.9855 & 0.0208 \\
\cmidrule(r){1-1} \cmidrule(r){2-4}
\cmidrule(r){5-7} \cmidrule(r){8-10}
\cmidrule(r){11-13}
     w./ Normal + ARAR Reg & \textbf{40.92} & \textbf{0.9948}  & \textbf{0.0139} & \textbf{32.97} & \textbf{0.9773} & 0.0277 & \textbf{29.45} & \textbf{0.9758} & \textbf{0.0360} & \textbf{35.49 }& \textbf{0.9864} & 0.0202 \\
    \bottomrule
    \multicolumn{1}{c}{\multirow{2}[2]{*}{Method}} &  \multicolumn{3}{c}{Standup}  & \multicolumn{3}{c}{Mutant} & \multicolumn{3}{c}{Trex} & \multicolumn{3}{c}{Lego}  \\
    \cmidrule(r){2-4} \cmidrule(r){5-7} 
    \cmidrule(r){8-10} \cmidrule(r){11-13}
        & PSNR↑ & SSIM↑ & LPIPS↓ & PSNR↑ & SSIM↑ & LPIPS↓ & PSNR↑ & SSIM↑ & LPIPS↓ & PSNR↑ & SSIM↑ & LPIPS↓  \\
\cmidrule(r){1-1} \cmidrule(r){2-4}
\cmidrule(r){5-7} \cmidrule(r){8-10}
\cmidrule(r){11-13}
    Baseline   & 38.11 & 0.9898  & \textbf{0.0138} & 37.59 & 0.9880 & \textbf{0.0167} & \textbf{34.23} & \textbf{0.9850} & \textbf{0.0131} & \textbf{25.03} & \textbf{0.9376} & \textbf{0.0382} \\
\cmidrule(r){1-1} \cmidrule(r){2-4}
\cmidrule(r){5-7} \cmidrule(r){8-10}
\cmidrule(r){11-13} 
    w./ Normal Reg   & 38.12 & 0.9898 & 0.0155 & 37.26 & 0.9864 & 0.0195 & 33.90 & 0.9841 & 0.0231 & 25.01 & 0.9371 & 0.0567  \\
\cmidrule(r){1-1} \cmidrule(r){2-4}
\cmidrule(r){5-7} \cmidrule(r){8-10}
\cmidrule(r){11-13}
     w./ Normal + ARAR Reg & \textbf{38.61} & \textbf{0.9903} & 0.0150 & \textbf{37.76}  & \textbf{0.9884} & 0.0191 & 34.21 & 0.9848 & 0.0230 & 25.02 & 0.9374 & 0.0583  \\
    \bottomrule
    \end{tabular}%
    }
  \label{table: ablation}%
\end{table*}%

\section{S3. Implementation details}
Given that each moment in the D-NeRF~\cite{pumarola2021d} dataset provides only a single viewpoint image, we constructed multiple virtual camera views at each moment to render the RGB images and corresponding depth maps for each training view. Then, we use the TSDF Fusion algorithm~\cite{newcombe2011kinectfusion} to extract mesh. In our paper, we use $10^{-3}$ as the default unit for the Chamfer Distance.

\end{document}